# Multi-Task Learning for Robot Perception with Imbalanced Data

Özgür Erkent[1]

[1]Hacettepe Üniversitesi, Mühendislik Fakültesi, Bilgisayar Mühendisliği, Ankara

**Abstract**

Multi-task problem solving has been shown to improve the accuracy of the individual tasks, which is an important feature for robots, as they have a limited resource. However, when the number of labels for each task is not equal, namely imbalanced data exist, a problem may arise due to insufficient number of samples, and labeling is not very easy for mobile robots in every environment. We propose a method that can learn tasks even in the absence of the ground truth labels for some of the tasks. We also provide a detailed analysis of the proposed method. An interesting finding is related to the interaction of the tasks. We show a methodology to find out which tasks can improve the performance of other tasks. We investigate this by training the teacher network with the task outputs such as depth as inputs. We further provide empirical evidence when trained with a small amount of data. We use semantic segmentation and depth estimation tasks on different datasets, NYUDv2 and Cityscapes.

## Dengesiz Verilerle Robot Algısı için Çok Görevli Öğrenme

**Öz**

Robotlarda, çoklu görev problemini çözmenin tekil görevlerin başarımını artırdığı gözlemlenmiştir. Bu, robotlar sınırlı bir kaynağa sahip olduklarından önemli bir niteliktir. Ancak her bir görev için etiket sayısı eşit olmadığında, yani dengesiz veri olduğunda, numune sayısının yetersiz olmasından dolayı sorun ortaya çıkmaktadır ve etiketleme farklı ortamlarda mobil robotlar için kolay değildir. Bu çalışmada, bazı görevler için doğruluk etiketlerinin eksik olduğu durumlarda bile görevlerin hepsini öğrenebilecek bir metot öneriyoruz. Ayrıca önerilen yöntemin ayrıntılı bir analizini de sunuyoruz. İlginç bir bulgu, görevlerin etkileşimiyle ilgilidir. Hangi görevlerin diğerlerinin performansını iyileştirebileceğini bulmak için bir metodoloji gösteriyoruz. Bunu, öğretici ağın eğitiminde girdi olarak derinlik gibi diğer görevlerin çıktılarını kullanarak araştırıyoruz. Ayrıca, az miktarda veriyle eğitildiğinde deneysel kanıt sağlıyoruz. NYUDv2 ve Cityscapes gibi farklı veri kümelerinde anlamsal bölütleme ve derinlik tahmini görevlerini kullanıyoruz.

**Anahtar Kelimeler:** otonom robotlar; bölütleme; derinlik tahmini; çok görevli öğrenmede dengesiz veriler

## Introduction

Autonomous robots need to achieve multiple perception tasks to be able to comprehend their surroundings and perform actions accordingly. In recent years, Deep Neural Networks (DNNs) that can learn multi-tasks simultaneously have been an important research topic (Kokkinos, 2017), (Crawshaw, 2020), (Vandenhende, et al., 2022). Studies have shown that multi-task learning (MTL) methods can achieve a performance close to single task methods; furthermore, they can surpass the performance under certain conditions (Kendall, Gal, & Cipolla, 2018), (Sener & Koltun, 2018), (Zamir, et al., 2018), (Standley, et al., 2020). However, MTL mainly relies on labeled data for training and to obtain the labeled data for all the tasks is cumbersome. In some cases, the labels for all the tasks may not be available. When the representations of the classes are not equal in a dataset, it is called as imbalanced (Chawla, Bowyer, Hall, & Kegelmeyer, 2002). Only a few studies have focused on this point (Ghiasi, Zoph, Cubuk, Le, & Lin, 2021), (Xu, Yang, & Zhang, 2023) in the MTL setting; however, their main focus was to train a teacher for missing task labels with uni-modal classifiers for each task without taking advantage of extra label information from task with excessive labels.

In this work, we propose a methodology to find out which tasks can improve the performance of other tasks via using a multi-modal teacher. We use the available ground truth labels of one of the tasks as input together with the original sensor input to the teacher network to produce the missing labels of the other task. This is used as pseudo-label to train the student network. We use a multi-modal network as a teacher since it will be capable of exploring more information to produce pseudo-labels for the student network which would result in higher accuracy; it should be noted that uni-modal methods were previously proposed in the literature (Ghiasi, Zoph, Cubuk, Le, & Lin, 2021); however, they are not capable of using all the available data. We analyze the conditions under which our assumption is valid, as it is known that multi-task cannot be successful for all tasks under all conditions (Standley, et al., 2020), (Vandenhende, et al., 2022), (Milli, Erkent, & Yılmaz, 2023). Unlike other studies, that require a vast amount of balanced data to find out the tasks that support each other, we develop a method to analyze the labels that can improve the performance of each other in

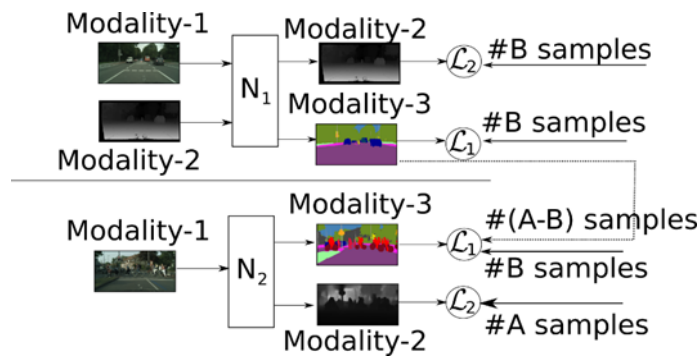

**Figure SEQ Figure \* ARABIC 1.** A sample multi-modal multi-task teacher DNN at the top () and a uni-modal multi-task student DNN at the bottom with two tasks (). Modality-1 is RGB image, Modality-2 (Mod2) is depth and Modality-3 (Mod3) is semantic segmentation in this example. and compute the losses for the tasks. We assume we have #A labeled samples for Mod2 and #B labeled samples for Mod3 where #A>#B for this example. First, the top DNN optimizes its parameters by using the #B samples. Then, #(A-B) pseudo labels are obtained by infering the top DNN and these labels are provided to compute the loss function of the ; therefore uses all samples efficiently to optimize its parameters

an MTL setting with a small and imbalanced number of samples.

We claim the following contributions:

- We show a methodology that uses one of the tasks as an input modality during training to find out the tasks that will improve the performance of the other tasks.

- We analyze the effect of pseudo labels in MTL produced by a multi-modal teacher to efficiently optimize the parameters,

- We provide empirical evidence for the effect of using pseudo-labels for imbalanced data with missing labels for various datasets,

An overview of our proposed method is shown in Figure 1. Here, $N_1$ is trained with B number of samples, while $N_2$ is trained with a number of samples, including pseudo-labels obtained from network $N_1$. We analyze the efficiency in Method Section and show the performance in Experiments Section. We will first discuss related works in Related Work. Finally, we summarize our work with a brief conclusion.

## Related Work

MTL has been studied in a different number of settings since it provides an improvement in performance, both in terms of accuracy and computation time. For a detailed survey, we refer the readers to recent surveys on this topic (Crawshaw, 2020), (Vandenhende, et al., 2022). An early study had shown that, a multi-task DNN, UberNet, can perform seven different tasks simultaneously and the performances of the tasks do not degrade (Kokkinos, 2017).

As multi-task networks improved the performance for certain tasks, datasets with multi-tasks, such as Taskonomy (Zamir, et al., 2018), have been collected. However, it has been shown that all the tasks do not support the performances of each other (Standley, et al., 2020). This is a critical and important empirical finding. A detailed theoretical background is not provided on how we can determine the tasks that support each other. In this study, we develop a methodology to find out which tasks improve the performance of each other without extensive training and without using large, labeled datasets.

Depending on their usage, different solutions have been developed for multi-task networks. Some methods use switching networks to select a certain task at a time; although they can achieve high performance (Sun, et al., 2021), their computation time and memory requirements are usually not within the reach of a mobile robot platform. We will consider networks which use layers with shared weights (generally a shared backbone) and perform tasks simultaneously, since the shared backbone provides an important advantage in terms of computation time and memory requirements. For example, (Rottmann, Posewsky, Milioto, Stachniss, & Behley, 2021) uses an initial MT network to boost the performance of depth estimation, then uses only depth labels to train a single task network. (Lee, Rameau, Im, & So, 2022) uses semantic priors to estimate the depth from a monocular camera. (He, Yang, Zhang, Long, & Song, 2024) considers balancing tasks in terms of their difficulty. Pad-Net (Xu, Ouyang, Wang, & Sebe, 2018) uses initial estimates as priors, then combines them via a multi-modal distillation module. A later work, called as MTI-NET (Vandenhende, Georgoulis, & Van Gool, 2020) again computes initial estimates; however, this time at different scales, and joins the features from these scales via a feature propagation module and finally combines these features again with a multi-modal distillation module. Although these methods may achieve high accuracies for depth estimation and semantic segmentation in MTL settings, they consider a balanced dataset for each task. However, it may not be always possible to label all the samples in a dataset, especially for mobile robot platforms, which results in imbalanced data. It should be noted that depth information can usually be obtained much more easily than the semantic segmentation labels in robotic studies.

Although not necessarily restricted to MTL setting, imbalanced data problems have been considered in previous studies. To overcome the problem of imbalanced data, data-level balancing has been used (Chawla, Bowyer, Hall, & Kegelmeyer, 2002); however, these methods don't use all the available data, therefore some labeled information may be lost. The loss-weight methods assign larger weights to the classes represented with minor samples. This method is also used in MTL tasks such as providing a weight of zero for losses of non-existing corresponding task samples (Leang, Sistu, Burger, Bursuc, & Yogamani, 2020); however, this does not always increase the accuracy of both tasks as will be shown in the experiments. We will finally mention generative adversarial networks (GANs) for imbalanced data problem, missing samples are generated by using GANs; however, the problem arises when the number of samples are not sufficient to create these synthetic samples (Wang, et al., 2022), (Choi, Jung, Kim, & Yoon, 2022). In our approach, we use the available samples to produce pseudo-labels; therefore, we don't require synthetic samples, and we also investigate cases where we don't have enough samples to produce synthetic data.

A few studies have tackled the problem by using teacher-student networks. For example, (Ghiasi, Zoph, Cubuk, Le, & Lin, 2021) have used uni-modal teachers to produce pseudo labels for the student. But, in this case, the extra information that is provided by the additional modality is not used efficiently. We provide a multi-modal approach to this problem where the extra labels are used to train the multi-modal network. An overview of the methods is provided in Table.1. Ours is the only method which considers imbalanced data in multi-task setting with sensor fusion for teacher training.

In the next Section, we provide a deep analysis for our method and analyze three Conditions. However, as it is intractable to find an upper bound for multi-task problem (Boursier, Konobeev, & Flammarion, 2022), we provide an empirical insight in Experiments Section.

**Table 1. Methods considering MT classification and imbalanced data labeling**

|  | MT | Imbalanced Data Consideration | Teacher Training with Sensor Fusion |
|---|---|---|---|
| (Vandenhendeet. Al., 2020) | + | - | - |
| (Chawla et. al., 2002) | - | + | - |
| (Leang, et. al., 2020) | + | - | - |
| (Wang, et al., 2022) | + | + | - |
| (Ghiasi et. al., 2021) | + | + | - |
| (Xu, Yang, & Zhang, 2023) | + | + | - |
| Ours | + | + | + |

## Method

Assume that we have two functions $f_1$ and $f_2$ that can generate outputs given input $z$, that is $X_1 = f_1(z)$ and $X_2 = f_2(z)$. Let us also assume that we have a third multi-task function that can output these two functions simultaneously $g(f_3(z)) = (f_1(z), f_2(z))$, where $f_3$ is another function, that can be considered as a function that is generating features (similar to layers in a DNN).

We have a set of labels $X_1^a, X_2^a$ for inputs $z^a$, $a \in A$; and $X_1^b$ for inputs $z^b$, $b \in B$. That is, we don't have labels for $X_2^b$. We will provide estimation functions $\hat{g}(f_3(z); \theta_g)$; and $\hat{f}_3(z; \theta_3)$ by using available data and labels where $\theta_g$ and $\theta_3$ are parameters to be optimized.

First, we can approximate the functions $f_1, f_2$ and $f_3$ as a sum of their sub-functions:

$$\widetilde{f}_a(z) \approx f_{a0}(z) + f_{a1}(z) + f_{a2}(z) + \cdots = \sum_{n=0}^{N} f_{an}(z) \qquad (1)$$

Note that $\left(f_a - \widetilde{f}_a\right) = 0$. Our main assumption is that the sub-functions of $f_3$ can be written as a linear combination of sub-functions of $f_1$ and $f_2$, as follows:

$$f_3(z) \approx \sum_{i \in I} f_{3i}(z) + \sum_{j \in J} f_{3j}(z) + \sum_{k \in K} f_{3k}(z)$$

$$f_{3i}(z) \approx \alpha_l f_{1l}(z) \approx \beta_m f_{2m}(z); \; l \in L, \; m \in M \qquad (2)$$

$$f_{3j}(z) \approx \alpha_o f_{1o}(z); \; o \in O \, f_{3k}(z) \approx \beta_p f_{2p}(z); \; p \in P$$

where $\alpha > 0$ and $\beta > 0$ are hyperparameters; $I, J, K$ are sub-function sets of $f_3$; $L, O$ are sub-function sets of $f_1$; and $M, P$ sub-function sets of $f_2$. That is, if $f_3$ is producing features, its features can be produced by using the features of functions (sub-functions of) $f_1$ and $f_2$.

We can approximate $g$ and $f_3$ by using a deep neural network $\hat{g}(f_3(z); \theta_g)$, $\hat{f}_3(z; \theta_3)$ with parameters $\theta_3$ and $\theta_g$ and these parameters can be obtained stochastically given sufficient number of labeled data samples $X_1$ and $X_2$. The output of $f_3$ can be similar to the features obtained in the network. Similarly, the sub-functions of $f_1$ and $f_2$ can also be features obtained in the network, if $f_1$ and $f_2$ functions were realized by neural networks.

We use a multi-modal multi-task estimator $h(X_1, z; \theta_h)$ to estimate the labels $X_1, X_2$ (teaching network). Note that its performance on $X_1$ is expected to be high since it will auto-encode the input $X_1$. The network will have a shared backbone for feature extraction for both tasks. Its parameters $\theta_h$ will be optimized on available labels $X_1^a, X_2^a$ and then the pseudo labels $\widetilde{X_2^b}$ will be estimated with this multi-modal estimator. Pseudo-labels are the outputs of the teacher network $h(X_1, z; \theta_h)$. The teacher network is a multi-modal input classifier network whose input is the excessive labels and output is the pseudo-label for the missing labels of the other tasks. Pseudo-labels are used as missing ground-truth labels for training the student network. The uni-modal estimator $\hat{g}(f_3(z); \theta_g)$ will use the labels $\{X_1^a, X_2^a, X_1^b, \widetilde{X_2^b}\}$ to optimize its weights. Now the dataset is balanced with additional pseudo-labels. If the functions $f_1$ and $f_2$ share similar sub-functions (features), then if we can learn the weights of one task, it could help to learn the weights for the other task. We can analyse the usage of a multi-modal teacher under the following three conditions.

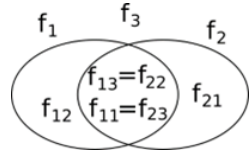

Figure 2. A sample illustration of Condition 1 for a very simple case.

*Condition 1:* $\left\| f_{3i} \right\| > \left\| f_{3j} \right\|$ and $\left\| f_{3i} \right\| > \left\| f_{3k} \right\|$, that is the number of shared sub-functions between tasks are high. Then the multi-modal teacher h will be able to optimize its parameters for the estimation of task $f_2$ since it will be able to learn the features for $f_1$ and $f_2$ simultaneously by using the labels from $X_1^a$ and $X_2^a$ in a balanced manner. Since most of them are common sub-functions (the features obtained from a large backbone in a neural network), then we will need to learn a smaller number of weights; or in other words, we will need less samples to learn the same number of weights. Then, the pseudo labels $\widetilde{X_2^b}$ will have a high estimation accuracy. note that, we would not have this advantage if we used a uni-modal teacher. A very basic sample illustration is shown in Figure 2. Here, the sub-functions are given as directly equal; however, they can be linear combinations as given in Eqn.2. $f_3$ has a total of 4 sub-functions, while $f_1$ and $f_2$ has both 3 sub-functions. In this

Figure, both tasks are expected to improve each other's performance since the number of common sub-functions, $\left\|f_{3i}\right\| > \left\|f_{3j}\right\| = \left\|f_{3k}\right\|$ are higher than uncommon ones. We have two more conditions to be considered.

*Condition 2:* $\left\|f_{3i}\right\| > \left\|f_{3j}\right\|$ and $\left\|f_{3k}\right\| > \left\|f_{3i}\right\|$, that is the number of sub-functions of $f_3$ has more similar sub-functions (features) with $f_2$ than $f_1$. In this case, to provide $\left(X_1, z\right)$ as input to the function $h$, that is $h\left(X_1, z\right)$, would not provide an increase in the performance of the estimation of pseudo-labels $\widetilde{X}_2$, since the features (sub-functions) to predict $X_1$ would be learned, but the majority of the features (sub-functions) to estimate $X_2$ would not be learned as $\left\|f_{3k}\right\| > \left\|f_{3i}\right\|$. On the other hand, if we provided $\left(X_2, z\right)$ as input to the function $h$, $h\left(X_2, z\right)$, would improve the estimation performance of the pseudo-labels $\widetilde{X}_1$, since the features (sub-functions) to predict $X_2$ would also constitute the majority of the features (sub-functions) to predict $X_2$ as $\left\|f_{3i}\right\| > \left\|f_{3j}\right\|$.

This condition also has a mirror case; we briefly mention that for $\left\|f_{3i}\right\| > \left\|f_{3k}\right\|$ and $\left\|f_{3j}\right\| > \left\|f_{3i}\right\|$, then input $\left(X_1, z\right)$ would improve the performance of $h$ to estimate $X_2$, but input $\left(X_2, z\right)$, that is the function $h\left(X_2, z\right)$, would not improve the performance to estimate $X_1$.

Therefore, we can conclude that if the sub-functions (features) of a task are in accordance with this condition, then one of the tasks will improve the performance of the other one, but the other task will not improve.

*Condition 3:* $\left\|f_{3j}\right\| > \left\|f_{3i}\right\|$ and $\left\|f_{3k}\right\| > \left\|f_{3i}\right\|$, in this condition, the functions of $f_1$ and $f_2$ have less common sub-functions than uncommon ones. Therefore, MT training would not increase the accuracy of these two functions. Multimodality would not provide accurate pseudo labels. This can be due to two main reasons.

The first reason may be that the number of provided samples would not be sufficient to find the proper approximations to the sub-functions of $f_1$ and $f_2$ stochastically. In this case, the similarities between these two modalities cannot be found, and the performance cannot be increased. The second reason may be that even if sufficient samples are provided, the sub-functions (features) of $f_1$ and $f_2$ may have a small number of similarities.

## Experiments

### Toy Example

First, we evaluate our method by using two simple functions for a better clarification,

$$f_1(z) = sin(z) + \sqrt{z}/300.0 + 2 + 3sin(5z)cos(2z)$$

$$f_2(z) = cos(z/3) + \sqrt{(z)}/200.0 + 1 - sin(3z) * cos(4z)$$

and we try to approximate $f_2$ by using a fully connected neural network with 4 layers containing 128 neurons each. If the network is single-task (ST), then it has single output value; if it is MT (Multi-Task), then it has two outputs, one for each function estimation, $f_1$ and $f_2$. The input variables (z) are taken in between [0,1], and we have 640 random variables for $f_1$ and 160 random variables for $f_2$. First, we train $f_2$ with 160 points on ST network, the results are shown in Figure 3. As can be seen, the results are not satisfactory with an error of $0.0149 \pm 0.0111$ as shown in Table.2. Next, we use the similar network with the modification of two outputs, one for $f_1$ and one for $f_2$. We train the network with 160 samples. We observe an improvement in the results. Finally, we train a Teacher MT Network (network $h(z, f_1)$) whose output is $f_2$ and $f_1$ with a=160 samples. Then, we create pseudo $\tilde{f}_2$ outputs by using 640 $f_1$ samples. We train the student single-task network by using these pseudo $\tilde{f}_2$ values. As it can be seen, this setting has the highest accuracy, with an approximate improvement of 30% with respect to MTL. It should be noted that the two functions are in accordance with Condition 1 and the results support our assumptions.

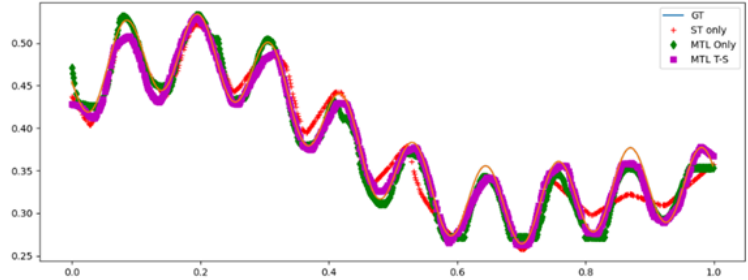

**Figure 3.** The results for the toy example. GT: Ground-truth. ST: Single-task. MTL T-S: Multi-task-learning with teacher-student approach. Best viewed in color.

**Table 2.** Avg MSE Errors for Toy Example

| ST | MTL Only | MTL T-S |
|---|---|---|
| $0.0149 \pm 0.0111$ | $0.0101 \pm 0.0068$ | $0.0072 \pm 0.0062$ |

**NYUDv2**

Next, we evaluate the multi-modal teacher, uni-modal student in NYUDv2 dataset (Nathan Silberman & Fergus, 2012). This is a dataset that consists of indoor images taken with an RGBD camera. It contains RGB images with corresponding depth and 40 semantic segmentation classes.

We use a multi-task DNN which uses distillation for semantic segmentation and depth estimation (Vandenhende, Georgoulis, & Van Gool, 2020) and we modify it by including a Feature Fusion Module (FFM) for multi-modality as shown in Figure 4. The details of the network are as follows. We use an HRNET (HRNet_w18) as **BB** to extract features from RGB images (Wang, et al., 2020) with pretrained weights from ImageNet for RGB. If an additional modality such as depth image or a semantically segmented image is used, then its features are extracted with another HRNet_w18. Learning rate is 0.0001 and batch size is 4. Note that additional modality is available only to the teacher network. **FFM** fuses the features from different modalities via adding them up at each scale level that is our contribution to the MT network for multi-modality.

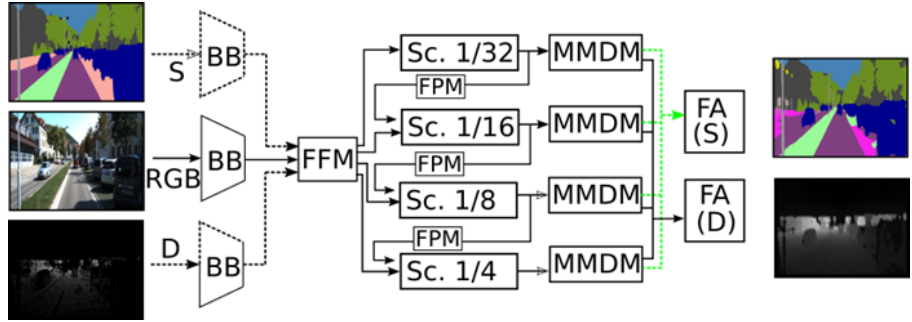

**Figure 4.** The network used in the experiments. BB: Backbone. FFM: Feature Fusion Module. FPM: Feature propagation module. MMDM: Multi-modal distillation module. FA: Feature Aggregation. The details of the modules are provided briefly in the text. The black dashed lines are used if the semantic segmented images or depth/Lidar images are given as input.

Feature Pyramid Network (FPN) is used to fuse features at four different scales (1/4, 1/8, 1/16, 1/32). The initial estimations at each scale are used as features and they are provided to larger scale estimations via **FPM**, which convolves and upscales the features. Finally, an **MMDM** is applied on different scales of the features and **FA** aggregates them by attending to the most relevant features to obtain the resulting task predictions. We use the following total loss for training

$$L = L_D + \alpha L_S$$

where $L_S$ denotes cross-entropy loss for semantic segmentation while $L_D$ denotes mean absolute loss for depth estimation. $\alpha$ is a parameter to provide a weight for balancing the losses between tasks. We have used $\alpha = 1$. The decision was made after a cross validation study on training data. We are using a custom number of training images than the original dataset, since we want to test our methodology which assumes imbalanced data. We use 500 depth images and 250 labeled images for semantic segmentation and 395 images for validation. This is a typical situation in robotics, where we have more depth images than semantically segmented images. Since the number of labels are much less than the original dataset, the results are also expected to have an inferior accuracy than the ones that are trained on the whole dataset. Initially, we performed an ablation study to see the effect of using pseudo-labels. For control, the input is a single RGB, output is depth and semantics (Uni-Modal-Multi-Task, UMMT). 500 depth and 250 segmentation labels are used for training UMMT. 250 semantic and depth labels are used to train the teacher to produce pseudo-labels (UMMT-PS). Then, 250 depth labels without corresponding semantic labels are used for obtaining 250 new pseudo-labels. These pseudo-labels together with original 500 depth and 250 semantic labels are used to train the student. The results are given Table.3. The ablation study reveals that pseudo-label method produces superior results.

**Table 3.** NYUDv2 Ablation Study.

|          | mIoU | rmse  | logrmse | ||f1|| | ||f2|| |
|----------|------|-------|---------|--------|--------|
| **UMMT**    | 22.1 | 0.790 | 0.264   | D      | S      |
| **UMMT-PS** | **25.3** | **0.653** | **0.206** | D      | S      |

**Table 4.** NYUDv2 Results.

|          | mIoU | Rmse  | logrmse | Inputs | Out   | loss | ||A|| | ||B|| | ||v|| | ||f1|| | ||f2|| |
|----------|------|-------|---------|--------|-------|------|-------|-------|-------|--------|--------|
| **MMST** | 24.0 | -     | -       | RGB + D | S    | GTS  | 250   | -     | 395   | S      | S      |
| **MMMT -D** | 24.4 | *0.337* | *0.102* | RGB + D | S + D | GTD  | 500   | 250   | 395   | D      | S      |
| **MMMT-S** | *52.6* | 0.805 | 0.261 | RGB + S | S + D | GTS + GTD | 500 | 250 | 395 | S | D |
| **FMST** | 19.7 | - | - | RGBD | S | GTS | 250 | - | 395 | S | S |
| **UMST-S** | 20.7 | - | - | RGB | S | GTS | 250 | - | 395 | S | S |
| **UMMT** | 22.1 | 0.790 | 0.264 | RGB | S + D | GTS + GTD | 250 | - | 395 | D | S |
| **UMMT-D** | 19.2 | 0.783 | 0.243 | RGB | S + D | GTD + GTS | 500 | 250 | 395 | D | S |
| **UMMT-PD** | 23.9 | 0.882 | 0.272 | RGB | S + D | GTD + PSD + GTS | 500 | 250 | 395 | S | D |
| **UMMT-PS** | **25.3** | **0.653** | **0.206** | RGB | S + D | PSS + GTS + GTD | 500 | 250 | 395 | D | S |

The results are shown in Table.4. The synonyms in the table are as follows: D: Depth, S: Semantic Segmentation, GTS: Ground Truth for S, GTD: Ground Truth for D, PSS: Pseudo-label for S, PSD: Pseudo-labels for D, MM: Multi-Modal, UM: Uni-Modal, ST: Single-Task, MT: Multi-Task. $\|A\|$ and $\|B\|$: number of labeled images for depth and semantic segmentation respectively (including pseudo labels). 295 validation images have been used for all cases. To make a comparison to the conditions in Methods Section, $f_1$ and $f_2$ are also provided. mIoU (mean intersection over union) is used for semantic segmentation measurement (the higher, the better) and root mean square (rmse) and logrmse are used for depth estimation measurement (the lower, the better).

First, we analyze the effect of multi-modality and multi-task on segmentation. The best results are shown in bold, while the results in italic mean that the output was used as input, therefore it is not included in comparison.

We use the same architecture as shown in Figure 4 for all the cases. Since we have a single estimation output in a single task network, we don't use any FA module in ST. The first row is a standard single segmentation task with input RGB (UMST-S).  In uni-modality, only RGB input branch is used, FFM is not used.

Next, we use the MT instead of ST with depth estimation and semantic segmentation tasks (UMMT-D). 500 depth labels and 250 segmentation labels are used. If the semantic segmentation has no label for the corresponding depth labels, $L_S$ is taken as zero, which is a standard procedure in MT studies (Leang, Sistu, Burger, Bursuc, & Yogamani, 2020). However, mIoU performance decreases from 20.7% to 19.23% with an rmse of 0.7825. This is due to overtraining of depth branch, which reduces the semantic segmentation performance.

Next, we use an MT, with equal number of labels, 250 from each task (UMMT). The performance of mIoU increases to 22.17%  and an rmse of 0.7896 is obtained. Therefore, it can be concluded that balanced data increases performance significantly. We also tested the effect of using multi-modality, when we use depth as additional input in an ST network, the accuracy increases to 24.02% (MMST). If we use multi-modal input with an MT network, the accuracy of mIoU further increases to 24.45% (MMMT-D). Although rmse has a high accuracy, it should be noted that depth is provided as input, this value only ensures that the data does not overfit. During runtime of the robot, we assume that depth will not be available. To be compliant with a robotic setting, we used MMMT-D as a teacher and used its outputs as pseudo-labels.

We used the labels which have a value of larger than a threshold of $\tau = 0.9$ on the last softmax layer of DNN. We trained a student network by using this balanced data which uses additional pseudo-labels for missing data (UMMT-PS). The mIoU increases to 25.38% and rmse reduces to 0.6534. These results show that our methodology that produces pseudo-labels via an MTL DNN increases the performance for NYUDv2 dataset.

This is in accordance with either Condition 1, that the sub-functions of $f_1$ (depth estimation) and $f_2$ (semantic segmentation) have more common sub-functions (features) than uncommon ones; or Condition 2 mirror situation, that is $f_1$ has more common sub-functions (features) with $f_3$ with respect to $f_2$. To be able to understand to which Condition the relation between semantic segmentation and depth estimation belongs, we perform the next evaluation.

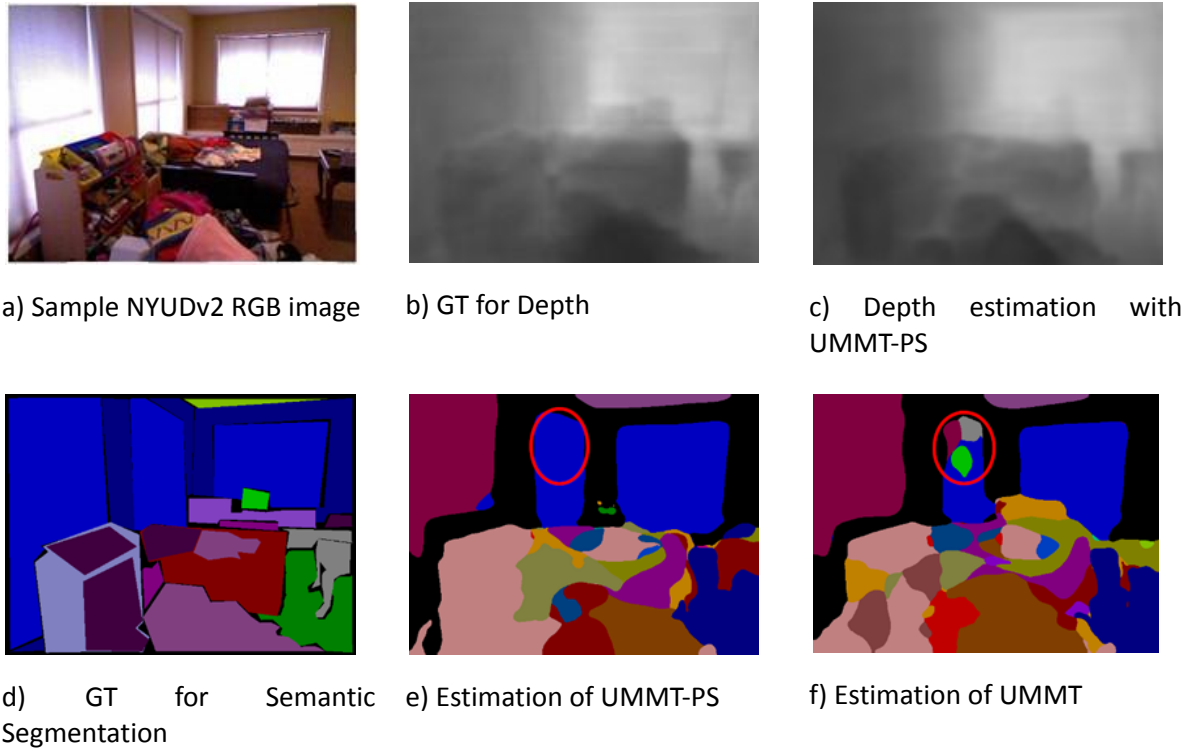

a) Sample NYUDv2 RGB image     b) GT for Depth          c) Depth estimation with UMMT-PS

d)   GT   for   Semantic     e) Estimation of UMMT-PS     f) Estimation of UMMT
Segmentation

Figure 5. Qualitative Results for a sample image from NYUDv2 dataset. The details are given in text.

We test if we can use a similar setting for obtaining pseudo-labels for depth. Here, we assumed that we have 250 labels for depth and 500 labels for semantic segmentation. Since multi-modal, multi-task (MMMT) setting had the best performance, we selected this setting to estimate the pseudo-depth labels. We have used semantic labels and RGB as input and obtained the missing pseudo-labels for depth (MMMT-S). We used these pseudo-labels to train a new network (UMMT-PD). However, as the results show, the performance did not improve.  This shows that the depth pseudo-labels were not correctly estimated, that is semantic segmentation does not improve depth segmentation. The results reveal that the relation between semantic segmentation and depth estimation belongs to Condition 2, that is the common sub-functions (features) of depth estimation provide a high estimation performance for semantic segmentation. But the sub-functions of semantic segmentation ($f_{3k}$ or $f_{2p}$ in Eqn.2) that are not common with depth estimation are higher ($\left\|f_{3k}\right\| > \left\|f_{3i}\right\|$).

Therefore, we can conclude that the usage of semantic segmentation labels as inputs do not improve the depth estimation in this case; but depth improves the semantic segmentation.

We show a sample qualitative result in Figure 5. On the first row, the RGB image and the corresponding depth ground truth (GT) and depth estimation is shown. The GT depth is noisy, which reduces the accuracy of the depth estimation. In the bottom, the semantic segmentation GT and the results are shown. In UMMT-PS, the area shown in circle is a single-class region, while UMMT wrongly estimates this region as multi-class segmentation. This sample illustrates a typical case for improvements by using pseudo-labels with a multi-task network.

**Cityscapes**

Next, we use Cityscapes (CS) dataset to evaluate our claims on a larger dataset taken with an autonomous vehicle. It contains 2975 images with depth and semantic segmentation information for training and 500 images for validation (Cordts, et al., 2016). The semantic labels have 19 classes. We have evaluated this dataset in two different configurations; first, we have used the large training set, 2975 images and 500 test images. Second, we have used a much smaller set of images, namely 300 for training and 200 for testing to test the effect of the number of samples.

We have used strategies which improved the performance similar to the NYUDv2 experiments and used the same DNN architectures. We used a learning rate of 0.0001 and a batch size of two. As the baseline, we have used UMST-S. Then, we used the multi-task setting UMMT, which improved the mIoU in the previous case. However, in this situation, the addition of the depth estimation task has reduced the segmentation performance, probably due to noisy labels of the depth. To be able to test the pseudo-labels, we have not used 250 labeled images which were used in in UMST-S; therefore, this drop in mIoU can be attributed to the reduction in the number of labeled images. Note that the depth estimation scores **rmse** and **logrmse** are higher (worse) with respect to NYUDv2. Since NYUDv2 was an indoor dataset, the distances were closer; whereas for CS, the data is collected outside and the distance ranges from a few meters to two hundred meters; therefore, the error rate is high for depth estimation. We provide the depth information as input to the network in MMMT-D, which will be the teacher for our final uni-modal network. In this case, mIoU increases to 73.6 % as expected; however, the depth estimation does not improve significantly. This is probably due to a higher learning rate for semantic segmentation. Finally, we use the additional 250 pseudo semantic labels estimated by MMMT-D and provide them as ground truth labels in loss computation by UMMT-PS. The semantic segmentation estimation and depth estimation by using RGB only improves the performance as shown in Table.5. mIoU for UMMT-PS achieves the second-best result in uni-modal estimation. The results have not improved as significantly as in NYUDv2 dataset, which may be attributed to the high number of samples.

The network learns its parameters and already provides a high performance due to large number of samples, and only a slight improvement of the performance may be possible for this DNN architecture. To test if this is the reason, we perform experiments with a limited number of labeled images. We use 300 semantic segmentation labels from the training set for uni-modal semantic segmentation task and evaluate on validation set as shown in UMST-S-L (L stands for **Low** number of samples). The performance is inferior to the model is trained with large number of samples, which is as expected. When we use a multi-task with depth estimation and semantic segmentation trained with 200 semantically segmented labels and 200 depth ground truths (UMMT-L), since we have a smaller number of semantically segmented labels, mIoU does not change much, which points to the fact that multi-task improves the performance of the network, and the performance does not degrade despite the smaller number of training samples for semantic segmentation. An interesting result is the improvement in the depth estimation results with respect to the model trained with a larger number of samples. This is probably due to the distribution of the depth samples in the smaller set. When we provide the depth as additional input to the model (MMMT-L), the semantic segmentation performance improves, but the depth estimation does not improve, which is again similar to the previous case when we provided large number of samples. Therefore, this can again be attributed to the difference in learning rates of semantic segmentation and depth estimation. Finally, again we use MMMT-L as the teaching network and use its outputs as the pseudo labels for semantic segmentation and train a new model called as UMMMT-PS-L whose input is RGB only. The performance of semantic segmentation improves significantly when we use a low number of images. Finally, the findings are in accordance with the Condition 2 found in the previous setup.

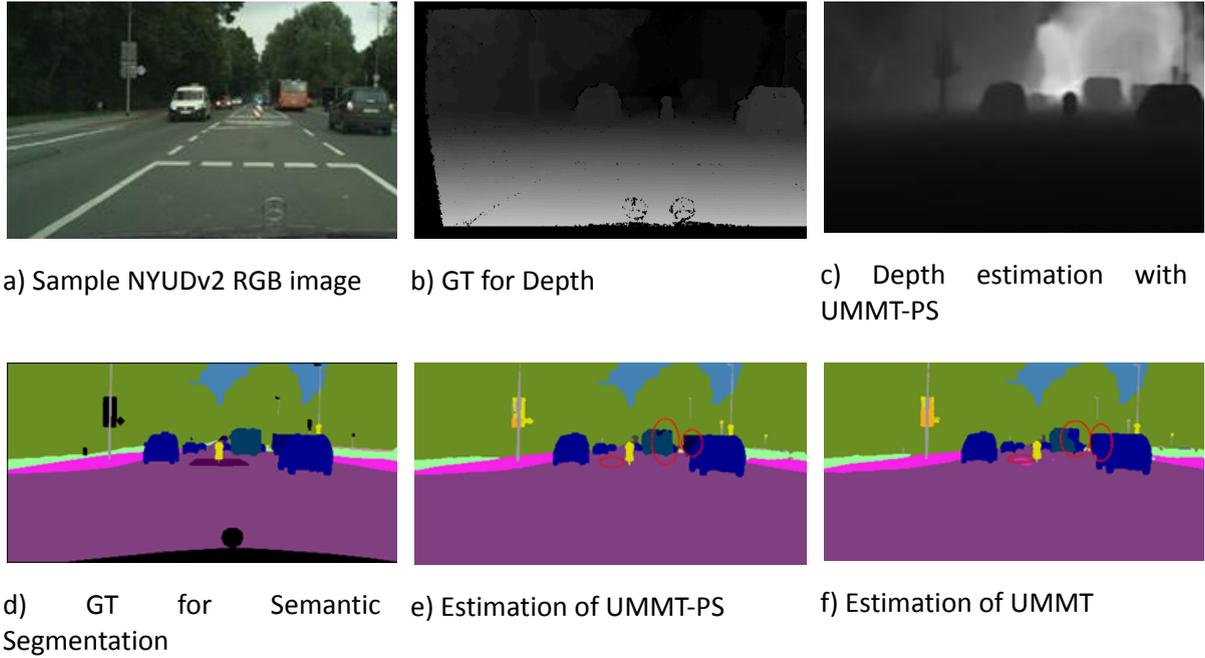

a) Sample NYUDv2 RGB image     b) GT for Depth          c) Depth estimation with UMMT-PS

d) GT for Semantic Segmentation     e) Estimation of UMMT-PS     f) Estimation of UMMT

**Figure 6.** Qualitative Results for a sample image from Cityscapes dataset. The details are given in text.

**Table 5.** Cityscapes Results.

|        | mIoU  | rmse  | logrmse | Inputs | Out   | loss            | \|\|A\|\| | \|\|B\|\| | \|\|v\|\| | \|\|f1\|\| | \|\|f2\|\| |
|--------|-------|-------|---------|--------|-------|-----------------|-----|-----|-----|-----|-----|
| UMMT   | 71.70 | 32.65 | **10.45** | RGB  | S + D | GTS + GTD       | 2975 | -   | 500 | D   | S   |
| MMMT-D | **73.60** | *33.80* | *10.40* | RGB + D | S + D | GTS + GTD    | 3475 | 2975 | 500 | D   | S   |
| UMST-S | 72.87 | -     | -       | RGB    | S     | GTS             | 2975 | -   | 500 | S   | S   |
| UMMT-PS | 73.10 | **32.60** | 10.50 | RGB   | S + D | PSS + GTS + GTD | 3475 | 2975 | 500 | D   | S   |
| UMMT-L | 52.64 | 27.92 | 9.56    | RGB    | S + D | GTS + GTD       | 300 | 200 | 200 | D   | S   |
| UMST-S-L | 52.83 | -   | -       | RGB    | S     | GTS             | 300 | 200 | 200 | S   | S   |
| MMMT-L | 53.84 | 27.70 | 9.68    | RGB+D  | S+D   |                 |     |     |     |     | S   |
| UMMT-PS-L | 54.06 | 29.11 | 7.84  | RGB    | S + D | PSS + GTS + GTD | 300 | 200 | 200 | D   | S   |

The qualitative results are shown in Figure 6. First, it should be mentioned that the depth obtained from disparity map has strong noise associated with it. The low accuracies on depth estimation are probably related to this issue. For the semantic segmentation, relevant improvements are shown with circled regions. For example, UMMT hallucinates a sideway in the middle of the road, while UMMT-PS does not. UMMT recognizes part of a bus as a car, while UMMT-PS does not. Finally, UMMT recognizes a truck behind a car as a car, while UMMT-PS correctly recognizes this as a truck.

**Conclusion**

We have proposed a methodology to use all the provided labels efficiently for a multi-task DNN in an imbalanced data setting. The usage of all data is important especially when the number of provided samples is small. We have proposed a method to produce pseudo-labels for the uni-modal multi-task network. It has been observed that, for a dual task including depth estimation and semantic segmentation, depth estimation can improve the semantic segmentation task via a multi-modal teacher network; however, the same is not valid for the semantic segmentation task. As part of the future work, the analyses of more than two tasks can be performed and the empirical evaluations related to several tasks can be performed.

**Author contribution**

The first author has developed the methods and performed the experiments.

**Etik**

Bu makalenin yayınlanmasıyla ilgili herhangi bir etik sorun bulunmamaktadır.

**Çıkar Çatışması**

Yazarlar herhangi bir çıkar çatışması olmadığını belirtmektedir.

**Acknowledgements:** This work was supported by TUBITAK under EU Commission Horizon 2020 Marie Skłodowska-Curie Actions Cofund program Circulation2 Scheme.

**ORCID**

*Birinci yazar Adı Soyadı* https://orcid.org/0000-0002-2436-3186